\documentclass[11pt,a4paper]{article}
\usepackage[hyperref]{emnlp-ijcnlp-2019}
\usepackage{times}
\usepackage{latexsym}

\usepackage{times}
\usepackage{latexsym}
\usepackage{hyperref}
\usepackage{url}
\usepackage{breakurl}
\usepackage{pifont}
\usepackage{verbatim}
\usepackage{booktabs}
\usepackage{amsmath}
\usepackage{amssymb}
\usepackage{xcolor}
\usepackage{graphics}
\usepackage{xspace}
\usepackage{float}
\usepackage{tikz}
\restylefloat{table}

\newcommand*\circled[1]{\tikz[baseline=(char.base)]{\node[shape=circle,draw,inner sep=1.4pt] (char) {#1};}}

\newcommand{\cmark}{\textcolor{green}{\ding{51}}}%
\newcommand{\xmark}{\textcolor{red}{\ding{55}}}%

\newcommand*\rot{\rotatebox{90}}
\usepackage{array,graphicx}

\newcommand{\tightparagraph}[1]{\noindent\textbf{#1:}}

\aclfinalcopy 

\title{An Evaluation Dataset for Intent Classification\\and Out-of-Scope Prediction}

\author{Stefan Larson ~~~~Anish Mahendran ~~~~Joseph J. Peper ~~~~Christopher Clarke\\ \textbf{Andrew Lee ~~~~Parker Hill ~~~~Jonathan K. Kummerfeld ~~~~Kevin Leach}\\
\textbf{Michael A. Laurenzano ~~~~Lingjia Tang ~~~~Jason Mars}\\
  Clinc, Inc.\\
  Ann Arbor, MI, USA \\
  {\tt stefan@clinc.com}
  \\}

\date{}

\begin{document}
\maketitle
\begin{abstract}
Task-oriented dialog systems need to know when a query falls outside their range of supported intents, but current text classification corpora only define label sets that cover every example.
We introduce a new dataset that includes queries that are out-of-scope---i.e., queries that do not fall into any of the system's supported intents. This poses a new challenge because models cannot assume that every query at inference time belongs to a system-supported intent class.
Our dataset also covers 150 intent classes over 10 domains, capturing the breadth that a production task-oriented agent must handle.
We evaluate a range of benchmark classifiers on our dataset along with several different out-of-scope identification schemes.
We find that while the classifiers perform well on in-scope intent classification, they struggle to identify out-of-scope queries.
Our dataset and evaluation fill an important gap in the field, offering a way of more rigorously and realistically benchmarking text classification in task-driven dialog systems.
\end{abstract}



\section{Introduction}
Task-oriented dialog systems have become ubiquitous, providing a means for billions of people to interact with computers using natural language. Moreover, the recent influx of platforms and tools such as Google's DialogFlow or Amazon's Lex for building and deploying such systems makes them even more accessible to various industries and demographics across the globe.

Tools for developing such systems start by guiding developers to collect training data for intent classification: the task of identifying which of a fixed set of actions the user wishes to take based on their query.
Relatively few public datasets exist for evaluating performance on this task, and those that do exist typically cover only a very small number of intents (e.g. \citet{snips}, which has 7 intents).
Furthermore, such resources do not facilitate analysis of \emph{out-of-scope} queries: queries that users may reasonably make, but fall outside of the scope of the system-supported intents.

\begin{figure}
    \centering
            \includegraphics[width=0.8\linewidth]{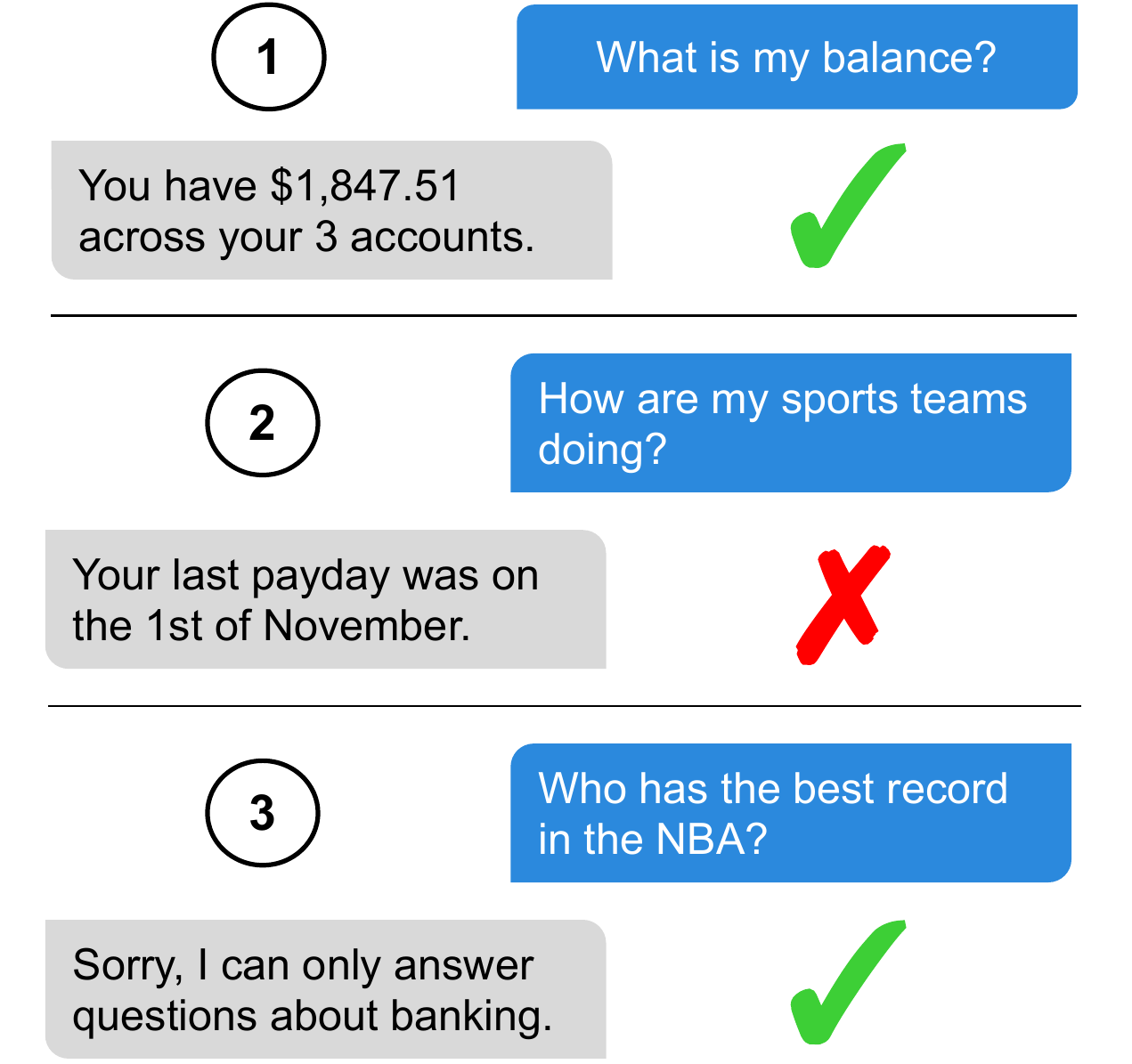}
    \caption{\label{fig:example}
    Example exchanges between a user (blue, right side) and a task-driven dialog system for personal finance (grey, left side). The system correctly identifies the user's query in \circled{1}, but in \circled{2} the user's query is mis-identified as in-scope, and the system gives an unrelated response. In \circled{3} the user's query is correctly identified as out-of-scope and the system gives a fallback response.
    }
\end{figure}

Figure \ref{fig:example} shows example query-response exchanges between a user and a task-driven dialog system for personal finance. In the first user-system exchange, the system correctly identifies the user's intent as an in-scope \textsc{balance} query. In the second and third exchanges, the user queries with out-of-scope inputs. In the second exchange, the system incorrectly identifies the query as in-scope and yields an unrelated response.
In the third exchange, the system correctly classifies the user's query as out-of-scope, and yields a fallback response.

Out-of-scope queries are inevitable for a task-oriented dialog system, as most users will not be fully cognizant of the system's capabilities, which are limited by the fixed number of intent classes. 
Correctly identifying out-of-scope cases is thus crucial in deployed systems---both to avoid performing the wrong action and also to identify potential future directions for development.  
However, this problem has seen little attention in analyses and evaluations of intent classification systems.

\begin{table*}[ht!]
\centering
  \scalebox{0.85}{
  \begin{tabular}{ l l l }
    \toprule
    \textbf{Domain} & \textbf{Intent} & \textbf{Query}\\
    \hline
    \textsc{banking} & \textsc{transfer} & \emph{move 100 dollars from my savings to my checking}\\
    \textsc{work} & \textsc{pto request} & \emph{let me know how to make a vacation request}\\
    \textsc{meta} & \textsc{change language} & \textit{switch the language setting over to german}\\
    \textsc{auto \& commute} & \textsc{distance} & \emph{tell the miles it will take to get to las vegas from san diego}\\
    \textsc{travel} & \textsc{travel suggestion} & \emph{what sites are there to see when in evans}\\
    \textsc{home} & \textsc{todo list update} & \textit{nuke all items on my todo list}\\
    \textsc{utility} & \textsc{text} & \emph{send a text to mom saying i'm on my way}\\
    \textsc{kitchen \& dining} & \textsc{food expiration} & \emph{is rice ok after 3 days in the refrigerator}\\
    \textsc{small talk} & \textsc{tell joke} & \emph{can you tell me a joke about politicians}\\
    \textsc{credit cards} & \textsc{rewards balance} & \emph{how high are the rewards on my discover card}\\
    \hline
    \textsc{out-of-scope} & \textsc{out-of-scope} & \textit{how are my sports teams doing}\\
    \textsc{out-of-scope} & \textsc{out-of-scope} & \textit{create a contact labeled mom}\\
    \textsc{out-of-scope} & \textsc{out-of-scope} & \textit{what's the extended zipcode for my address}\\
    \bottomrule
  \end{tabular}
  }
\caption{\label{samples}
Sample queries from our dataset. The out-of-scope queries are similar in style to the in-scope queries.
}
\end{table*}

This paper fills this gap by analyzing intent classification performance with a focus on out-of-scope handling.
To do so, we constructed a new dataset with 23,700 queries that are short and unstructured, in the same style made by real users of task-oriented systems.
The queries cover 150 intents, plus out-of-scope queries that do not fall within any of the 150 in-scope intents.

We evaluate a range of benchmark classifiers and out-of-scope handling methods on our dataset.
BERT~\citep{bert} yields the best in-scope accuracy, scoring 96\% or above even when we limit the training data or introduce class imbalance. 
However, all methods struggle with identifying out-of-scope queries.
Even when a large number of out-of-scope examples are provided for training, there is a major performance gap, with the best system scoring 66\% out-of-scope recall.
Our results show that while current models work on known classes, they have difficulty on out-of-scope queries, particularly when data is not plentiful.
This dataset will enable future work to address this key gap in the research and development of dialog systems. All data introduced in this paper can be found at \url{https://github.com/clinc/oos-eval}.

\section{Dataset}

We introduce a new crowdsourced dataset of 23,700 queries, including 22,500 in-scope queries covering 150 intents, which can be grouped into 10 general domains. The dataset also includes 1,200 out-of-scope queries.
Table \ref{samples} shows examples of the data.\footnote{See the supplementary material for a full list of domains and intents, as well as additional examples.}

\subsection{In-Scope Data Collection}
We defined the intents with guidance from queries collected using a \emph{scoping} crowdsourcing task, which prompted crowd workers to provide questions and commands related to topic domains in the manner they would interact with an artificially intelligent assistant. We manually grouped data generated by \emph{scoping} tasks into intents. To collect additional data for each intent, we used the \emph{rephrase} and \emph{scenario} crowdsourcing tasks proposed by \citet{naacl18data}.\footnote{
  In all cases, crowd workers were paid \$0.20 per task.
  See the supplementary material for examples of each task.
}
For each intent, there are 100 training queries, which is representative of what a team with a limited budget could gather while developing a task-driven dialog system. Along with the 100 training queries, there are 20 validation and 30 testing queries per intent.

\subsection{Out-of-Scope Data Collection}
Out-of-scope queries were collected in two ways.
First, using worker mistakes: queries written for one of the 150 intents that did not actually match any of the intents.
Second, using \emph{scoping} and \emph{scenario} tasks with prompts based on topic areas found on Quora, Wikipedia, and elsewhere.
To help ensure the richness of this additional out-of-scope data, each of these task prompts contributed to at most four queries.
Since we use the same crowdsourcing method for collecting out-of-scope data, these queries are similar in style to their in-scope counterparts.

The out-of-scope data is difficult to collect, requiring expert knowledge of the in-scope intents to painstakingly ensure that no out-of-scope query sample is mistakenly labeled as in-scope (and vice versa).
Indeed, roughly only 69\% of queries collected with prompts targeting out-of-scope yielded out-of-scope queries.
Of the 1,200 out-of-scope queries collected, 100 are used for validation and 100 are used for training, leaving 1,000 for testing.

\subsection{Data Preprocessing and Partitioning}
For all queries collected, all tokens were down-cased, and all end-of-sentence punctuation was removed.
Additionally, all duplicate queries were removed and replaced.

In an effort to reduce bias in the in-scope data, we placed all queries from a given crowd worker in a single split (train, validation, or test). This avoids the potential issue of similar queries from a crowd worker ending up in both the train and test sets, for instance, which would make the train and test distributions unrealistically similar. We note that this is a recommendation from concurrent work by \citet{geva-2019}. We also used this procedure for the out-of-scope set, except that we split the data into train/validation/test based on task prompt instead of worker.

\subsection{Dataset Variants}
In addition to the full dataset, we consider three variations.
First, \textbf{Small}, in which there are only 50 training queries per each in-scope intent, rather than 100.
Second, \textbf{Imbalanced}, in which intents have either 25, 50, 75, or 100 training queries.
Third, \textbf{OOS+}, in which there are 250 out-of-scope training examples, rather than 100.
These are intended to represent production scenarios where data may be in limited or uneven supply.

\section{Benchmark Evaluation}

To quantify the challenges that our new dataset presents, we evaluated the performance of a range of classifier models\footnote{See the supplementary material for model details.}
and out-of-scope prediction schemes.

\subsection{Classifier Models}

\tightparagraph{SVM}
A linear support vector machine with bag-of-words sentence representations.
%
%

\tightparagraph{MLP}
A multi-layer perceptron with USE embeddings \citep{USE} as input.

\tightparagraph{FastText}
A shallow neural network that averages embeddings of \textit{n}-grams \citep{fasttext}.

\tightparagraph{CNN}
A convolutional neural network with non-static word embeddings initialized with GloVe \citep{glove}.


\tightparagraph{BERT}
A neural network that is trained to predict elided words in text and then fine-tuned on our data \citep{bert}.

\tightparagraph{Platforms}
Several platforms exist for the development of task-oriented agents.
We consider Google's DialogFlow\footnote{\url{https://dialogflow.com}} and Rasa NLU\footnote{\url{https://github.com/RasaHQ/rasa}} with spacy-sklearn.

\subsection{Out-of-Scope Prediction}
We use three baseline approaches for the task of predicting whether a query is out-of-scope: (1) \textbf{\emph{oos-train}}, where we train an additional (i.e. $151^{st}$) intent on out-of-scope training data; (2) \textbf{\emph{oos-threshold}}, where we use a threshold on the classifier's probability estimate; and (3) \textbf{\emph{oos-binary}}, a two-stage process where we first classify a query as in- or out-of-scope, then classify it into one of the 150 intents if classified as in-scope. 

To reduce the severity of the class imbalance between in-scope versus out-of-scope query samples (i.e., 15,000 versus 250 queries for OOS+), we investigate two strategies when using \emph{oos-binary}: one where we undersample the in-scope data and train using 1,000 in-scope queries sampled evenly across all intents (versus 250 out-of-scope), and another where we augment the 250 OOS+ out-of-scope training queries with 14,750 sentences sampled from Wikipedia.

From a development point of view, the \emph{oos-train} and \emph{oos-binary} methods both require careful curation of an out-of-scope training set, and this set can be tailored to individual systems. The \emph{oos-threshold} method is a more general decision rule that can be applied to any model that produces a probability. In our evaluation, the out-of-scope threshold was chosen to be the value which yielded the highest validation score across all intents, treating out-of-scope as its own intent.

\begin{table*}
\centering
 \scalebox{0.85}{
   \begin{tabular}{ cl c| rrrr c| rrrr}
    
    \toprule
               & & & \multicolumn{4}{c}{~~~~~~\textbf{In-Scope Accuracy}} & & \multicolumn{4}{c}{\textbf{Out-Of-Scope Recall}} \\
    & \textbf{Classifier} & & \textbf{Full} & \textbf{Small} & \textbf{Imbal} & \textbf{OOS+} & & \textbf{Full} & \textbf{Small} & \textbf{Imbal} & \textbf{OOS+} \\
    \midrule    
    & FastText   & & 89.0 & 84.5 & 87.2 & 89.2 & & 9.7 & 23.2 & 12.2 &  32.2 \\
    & SVM    & & 91.0 & 89.6 & 89.9 & 90.1 & & 14.5 & 18.6 & 16.0 & 29.8 \\
    & CNN        & & 91.2 & 88.9 & 89.1 & 91.0 & & 18.9 & 22.2 & 19.0 & 34.2 \\
    & DialogFlow & & 91.7 & 89.4 & 90.7 & 91.7 & & 14.0 & 14.1 & 15.3 & 28.5 \\
    & Rasa & & 91.5 & 88.9 & 89.2 & 90.9 & & 45.3 & \textbf{55.0} & \textbf{49.6} & \textbf{66.0} \\
    \rot{\rlap{~\textbf{~~~~\emph{oos-train}}}}& MLP & & 93.5 & 91.5 & 92.5 & 94.1 & & \textbf{47.4} & 52.2 & 35.6 &  53.9\\
    
    & BERT       & & \textbf{96.9} & \textbf{96.4} & \textbf{96.3} & \textbf{96.7} & & 40.3 & 40.9 & 43.8 & 59.2 \\
    
    \midrule
    \midrule
    & SVM  & & 88.2 & 85.6 & 86.0 & --- & & 18.0 & 13.0 & 0.0 & ---\\
    & FastText   & & 88.6 & 84.8 & 86.6 & --- & & 28.3 & 6.0 & 33.2 &  --- \\
    & DialogFlow  & & 90.8 & 89.2 & 89.2 & --- & & 26.7 & 20.5 & 38.1 & ---\\
    & Rasa  & & 90.9 & 89.6 & 89.4 & --- & & 31.2 & 1.0 & 0.0 & ---\\
    & CNN        & & 90.9 & 88.9 & 90.0 & --- & & 30.9 & 25.5 & 26.9 & --- \\
    \rot{\rlap{~\textbf{\emph{oos-threshold}}}}& MLP        & & 93.4 & 91.3 & 92.5 & --- & & 49.1 & 32.4 & 13.3 &  ---\\
    & BERT       & & \textbf{96.2} &  \textbf{96.2} & \textbf{95.9} & --- & & \textbf{52.3} & \textbf{58.9} & \textbf{52.8} & --- \\
    
    \bottomrule
  \end{tabular}
  }
\caption{\label{tab:classifier-results}
Benchmark classifier results under each data condition using the \emph{oos-train} (top half) and \emph{oos-threshold} (bottom half) prediction methods.
}
\end{table*}

\subsection{Metrics}

We consider two performance metrics for all scenarios:
(1) accuracy over the 150 intents, and
(2) recall on out-of-scope queries.
We use recall to evaluate out-of-scope since we are more interested in cases where such queries are predicted as in-scope, as this
would mean a system gives the user a response that is completely wrong. Precision errors are less problematic as the fallback response will prompt the user to try again, or inform the user of the system's scope of supported domains.

\section{Results}
\subsection{Results with \emph{oos-train}}
Table~\ref{tab:classifier-results} presents results for all models across the four variations of the dataset.
First, BERT is consistently the best approach for in-scope, followed by MLP. Second, out-of-scope query performance is much lower than in-scope across all methods. Training on less data (Small and Imbalanced) yields models that perform slightly worse on in-scope queries.
The trend is mostly the opposite when evaluating out-of-scope, where recall increases under the Small and Imbalanced training conditions. Under these two conditions, the size of the in-scope training set was decreased, while the number of out-of-scope training queries remained constant. This indicates that out-of-scope performance can be increased by increasing the relative number of out-of-scope training queries. We do just that in the OOS+ setting---where the models were trained on the full training set as well as 150 additional out-of-scope queries---and see that performance on out-of-scope increases substantially, yet still remains low relative to in-scope accuracy.

\subsection{Results with \emph{oos-threshold}}
In-scope accuracy using the \emph{oos-threshold} approach is largely comparable to \emph{oos-train}. Out-of-scope recall tends to be much higher on Full, but several models suffer greatly on the limited datasets. BERT and MLP are the top \emph{oos-threshold} performers, and for several models the threshold approach provided erratic results, particularly FastText and Rasa.

\subsection{Results with \emph{oos-binary}}
Table~\ref{binary} compares classifier performance using the \emph{oos-binary} scheme. In-scope accuracy suffers for all models using the undersampling scheme when compared to training on the full dataset using the \emph{oos-train} and \emph{oos-threshold} approaches shown in Table~\ref{tab:classifier-results}. However, out-of-scope recall improves compared to \emph{oos-train} on Full but not OOS+. Augmenting the out-of-scope training set appears to help improve both in-scope and out-of-scope performance compared to undersampling, but out-of-scope performance remains weak.

\begin{table}[t]
\begin{center}
\scalebox{0.85}{
\begin{tabular}{ c c c c c } 
 \toprule
 & \multicolumn{2}{c}{\textbf{In-Scope}} & \multicolumn{2}{c}{\textbf{Out-of-Scope}} \\
 & \multicolumn{2}{c}{\textbf{Accuracy}} & \multicolumn{2}{c}{\textbf{Recall}} \\
 \textbf{Classifier} & \textbf{under} & \textbf{wiki aug} & \textbf{under} & \textbf{wiki aug} \\
 \hline
 DialogFlow & 84.7 & --- & 37.3 & --- \\
 Rasa & 87.5 & ---  & 37.7 & --- \\
 FastText & 88.1 & 87.0 & 22.7 & 31.4 \\
 SVM & 88.4 & 89.3 & 32.2 & 37.7 \\
 CNN & 89.8 & 90.1 & 25.6 & 39.7 \\
 MLP & 90.1 & 92.9 & \textbf{52.8} & 32.4 \\
 BERT & \textbf{94.4} & \textbf{96.0} & 46.5 & \textbf{40.4} \\
 \hline
\end{tabular}}
\end{center}
\caption{\label{binary}
Results of \emph{oos-binary} experiments on OOS+, where we compare performance of undersampling (under) and augmentation using sentences from Wikipedia (wiki aug). The wiki aug approach was too large for the DialogFlow and Rasa classifiers.
}
\end{table}

\section{Prior Work}
\begin{table*}
\centering
\scalebox{1}{
\scalebox{0.85}{
  \begin{tabular}{ l  r  r  c  c  c  c }
    \toprule
            & \textbf{Num.} & \textbf{Num.}  & \textbf{Chatbot} & \textbf{Many}    & \textbf{Constrained}       & \textbf{Out-of-Scope} \\
    \textbf{Dataset} & \textbf{Intents} & \textbf{Utterances} & \textbf{Style}   & \textbf{Intents} & \textbf{Training Data} & \textbf{Utterances}    \\
    \midrule
    Our Dataset (This Work) & 150 & 23,700 & \cmark & \cmark & \cmark & \cmark \\
    TREC-6, \citep{TREC} & 6 & 5,952 & \xmark & \xmark & \xmark & \xmark \\
    TREC-50, \citep{TREC} & 50 & 5,952 & \xmark & \cmark & \cmark & \xmark \\
    Web Apps, \citep{cbc} & 8 & 89 & \xmark & \xmark & \cmark & \xmark \\
    Ask Ubuntu, \citep{cbc} & 5 & 162 & \xmark & \xmark & \cmark & \xmark \\
    Chatbot Corpus, \citep{cbc} & 2 & 206 & \cmark & \xmark & \cmark & \xmark \\
    Snips, \citep{snips} & 7 & 14,484 & \cmark & \xmark & \xmark & \xmark \\
    \citet{sds-eval} & 54 & 25,716 & \cmark & \cmark & \xmark & \xmark\\
    \bottomrule
  \end{tabular}
}}
\caption{\label{tab:data}
Classification dataset properties.
Ours has the broadest range of intents and specially collected out-of-scope queries. We consider ``chatbot style" queries to be short, possibly unstructured questions and commands.
}
\end{table*}
In most other analyses and datasets, the idea of out-of-scope data is not considered, and instead the output classes are intended to cover all possible queries (e.g., TREC \citep{TREC}).
Recent work by \citet{hendrycks17baseline} considers a similar problem they call out-of-distribution detection.
They use other datasets or classes excluded during training to form the out-of-distribution samples.
This means that the out-of-scope samples are from a small set of coherent classes that differ substantially from the in-distribution samples.
Similar experiments were conducted for evaluating unknown intent discovery models in \citet{lin-xu-2019-deep}.
In contrast, our out-of-scope queries cover a broad range of phenomena and are similar in style and often similar in topic to in-scope queries, representing things a user might say given partial knowledge of the capabilities of a system.

Table~\ref{tab:data} compares our dataset with other short-query intent classification datasets.
The Snips \citep{snips} dataset and the dataset presented in \citet{sds-eval} are the most similar to the in-scope part of our work, with the same type of conversational agent requests.
Like our work, both of these datasets were bootstrapped using crowdsourcing.
However, the Snips dataset has only a small number of intents and an enormous number of examples of each.
Snips does present a low-data variation, with 70 training queries per intent, in which performance drops slightly.
The dataset presented in \citet{sds-eval} has a large number of intent classes, yet also contains a wide range of samples per intent class (ranging from 24 to 5,981 queries per intent, and so is not constrained in all cases).

\citet{cbc} created datasets with constrained training data, but with very few intents, presenting a very different type of challenge.
We also include the TREC query classification datasets \citep{TREC}, which have a large set of labels, but they describe the desired response type (e.g., distance, city, abbreviation) rather than the action intents we consider.
Moreover, TREC contains only questions and no commands.
Crucially, none of the other datasets summarized in Table \ref{tab:data} offer a feasible way to evaluate out-of-scope performance.

The Dialog State Tracking Challenge (DSTC) datasets are another related resource.
Specifically, DSTC 1 \cite{dstc1}, DSTC 2 \cite{dstc2}, and DSTC 3 \cite{dstc3} contain ``chatbot style" queries, but the datasets are focused on state tracking. Moreover, most if not all queries in these datasets are in-scope.  In contrast, the focus of our analysis is on both in- and out-of-scope queries that challenge a virtual assistant to determine whether it can provide an acceptable response.

%

\section{Conclusion}
This paper analyzed intent classification and out-of-scope prediction methods with a new dataset consisting of carefully collected out-of-scope data. Our findings indicate that certain models like BERT perform better on in-scope classification, but all methods investigated struggle with identifying out-of-scope queries. Models that incorporate more out-of-scope training data tend to improve on out-of-scope performance, yet such data is expensive and difficult to generate. We believe our analysis and dataset will lead to developing better, more robust dialog systems.

All datasets introduced in this paper can be found at \url{https://github.com/clinc/oos-eval}.




\bibliographystyle{acl_natbib}
\end{document}